\pdfoutput=1

\documentclass[11pt]{article}

\usepackage{acl}

\usepackage{times}
\usepackage{latexsym}

\usepackage[T1]{fontenc}

\usepackage[utf8]{inputenc}

\usepackage{microtype}

\usepackage{inconsolata}

\usepackage{graphicx}

\usepackage[normalem]{ulem}  
\usepackage{xcolor}

\usepackage{diagbox}
\usepackage{booktabs}
\usepackage{multirow}
\usepackage{amsmath}
\usepackage{amssymb}
\usepackage{subcaption}
\usepackage{enumitem}
\usepackage{multicol}
\newcommand\mC[1]{\multicolumn{1}{c}{#1}}
\newcommand{\lexeme}[2]{\textsc{#1}::\texttt{#2}}

\title{Meta-Learning Neural Mechanisms rather than Bayesian Priors}

\author{
  \textbf{Michael Eric Goodale\textsuperscript{$\iota\epsilon$}},
  \textbf{Salvador Mascarenhas\textsuperscript{$\iota\epsilon$}},
  \textbf{Yair Lakretz\textsuperscript{$\lambda\epsilon$}},
\\
  \textsuperscript{$\iota$}Institut Jean Nicod \\
  \textsuperscript{$\lambda$}Laboratoire de Sciences Cognitives et Psycholinguistique\\
  \textsuperscript{$\epsilon$}Département d’études cognitives, ENS, EHESS, CNRS, PSL University.\\
  \texttt{
    {\{michael.goodale,salvador.mascarenhas\}@ens.fr}, yair.lakretz@gmail.com
  }
}

\begin{document}
\maketitle
\begin{abstract}
Children acquire language despite being exposed to several orders of magnitude less data than large language models require.
Meta-learning has been proposed as a way to integrate human-like learning biases into neural-network architectures, combining both the structured generalizations of symbolic models with the scalability of neural-network models.
But what does meta-learning exactly imbue the model with?
We investigate the meta-learning of formal languages and find that, contrary to previous claims, meta-trained models are not learning simplicity-based priors when meta-trained on datasets organised around simplicity.
Rather, we find evidence that meta-training imprints neural mechanisms (such as counters) into the model, which function like cognitive primitives for the network on downstream tasks.
Most surprisingly, we find that meta-training on a \textit{single} formal language can provide as much improvement to a model as meta-training on 5000 different formal languages, provided that the formal language incentivizes the learning of useful neural mechanisms.
Taken together, our findings provide practical implications for efficient meta-learning paradigms and new theoretical insights into linking symbolic theories and neural mechanisms.
\end{abstract}

\section{Introduction}

In the past decade, enormous advances in neural network models have led to a novel neural-network orthodoxy perhaps best exemplified by Richard Sutton's Bitter Lesson \citep{sutton_bitter_2019}.
The now prevailing view has been to eschew purpose-built architectures in favour of scale in larger models and larger training datasets rather than narrow purpose-built architectures.
Nevertheless, humans' capacity to generalize from very few examples has remained an elusive and intriguing goal for both AI practitioners and cognitive scientists sympathetic to connectionism.

\citet{yang_one_2022} introduced a symbolic model to capture aspects of humans' basic generalisations from small data on simple formal languages.
They use a ``language of thought'' model which combines simple cognitive ``primitives'' (e.g.\ simple functions like logical AND, getting the first element of a list, and so on) into symbolic programs explored via Bayesian-inference techniques.
The model can then ``learn'', given some data, by finding a program which generates the underlying data, potentially generalizing correctly into out-of-distribution data.
Crucially, the model has a Bayesian prior for simpler (e.g.\ shorter) programs.
The model can only learn effectively with toy formal grammars, for example it perfectly generalizes from the minuscule dataset, $\{a, aa, aaa\}$, to the formal grammar $a^n$. 
Despite the model's performance on simple grammars, its symbolic nature makes it difficult to scale up to more complicated material such as natural language, because the non-differentiable search over programs is intractable and cannot exploit gradient-descent.

One potential way to square the circle of symbolic modeling's theoretically-motivated generalizations with neural-network approaches' scaleable learning is meta-learning.
Meta-learning is an increasingly popular approach among cognitively-minded researchers to address humans' capacity for systematic generalization while staying within the familiar realm of mainstream neural-network techniques~\citep{lake_human-like_2023,binz_meta-learned_2024,mccoy_modeling_2023,mccoy_universal_2020}.
This approach has models \emph{learn} baseline expectations, in the form of ``Bayesian priors'' or ``inductive biases,'' approximating and hopefully reproducing the language-readiness of the human child's brain.
Meta-learning approaches thus avoid the challenges of selecting and hard coding an appropriate \emph{a priori} architecture, while leveraging the remarkable learning abilities of neural networks.
Consequently, several cognitive scientists have proposed that meta-learning of one form or another may provide solutions to the problems of out-of-distribution generalization on the basis of limited data \citep{lake_human-like_2023,binz_meta-learned_2024}.

\citet{mccoy_modeling_2023} use meta-learning in the special case of language learning: they use meta-learning techniques to, as they put it, ``distill'' a Bayesian prior for a particular kind of simplicity into a neural network.
They use model-agnostic meta-learning \citep[MAML;][]{finn_model-agnostic_2017} to meta-train LSTMs to learn simple formal languages from small amounts of data in the same schema as \citet{yang_one_2022}.
Their model is meta-trained on formal languages sampled with preference for simplicity, and then trains on datasets like those from \citet{yang_one_2022}.

Their model shows impressive improvements in performance compared to an un-metatrained neural network, with generalizations from small data comparable to \citeposs{yang_one_2022} symbolic model.
They describe their model as distilling a Bayesian prior for simplicity into a connectionist model, thus benefiting from the flexibility provided by neural networks.
In other words, they claim the model learns to mimic its meta-training distribution when trained on later tasks, thereby following the simplicity prior of its meta-training dataset.

Here we examine \citeposs{mccoy_modeling_2023} proposal for meta-learning in the case of formal languages in detail, and demonstrate that their conclusion is premature.
We provide an alternative \textit{mechanistic view}, which, in contrast to the \textit{simplicity-bias view}, suggests that meta-training encourages models to learn useful neural mechanisms (e.g.\ counters) and that meta-learning does not distill simplicity biases. 
The neural mechanisms then function as available ``neural primitives'' during task acquisition. 

We contrast the mechanistic and simplicity-bias views by testing each of their predictions, as spelled out in Table~\ref{tab:pred}, and find converging evidence in support of the mechanistic view, but not the simplicity-bias one: First, models can achieve the same behavioral outcomes using meta-learning datasets that in no way follow a Bayesian simplicity prior, in contrast to the prediction of the simplicity-bias view. 
Crucially, second, we find that \emph{a single formal language} can serve as a sufficiently rich meta-learning dataset in certain cases. 
These can be explained by the mechanistic view but not the simplicity-bias view.
Finally, manipulating the types of mechanisms a neural architecture can develop (e.g., preventing counting by replacing LSTMs by GRU units), we show that meta-learning effects disappear, as predicted by the mechanistic but not the simplicity-bias view.



\begin{table*}[htb]
  \begin{center}
  \begin{tabular}{p{7cm}p{7cm}}
    \toprule 
    \mC{\textbf{Simplicity-bias view}} & \mC{\textbf{Mechanistic-complexity view}}\\
  Meta-trained models learn to learn with a Bayesian simplicity prior.
  & 
   Meta-trained models learn useful neural-mechanisms.\\
   \midrule

   \begin{minipage}{7cm}
     \begin{enumerate}[leftmargin=*, label={1\alph*)}, ref={1\alph*)}]
       \item Simplicity-focus datasets will help the model learn to make the simplest generalization in later training. 
       \item Datasets without a prior for simplicity will make incorrect generalizations and perform worse. 
    \end{enumerate}
  \end{minipage} &
   \begin{minipage}{7cm}
    \begin{enumerate}[leftmargin=*, label={2\alph*)}, ref={2\alph*)}]
       \item For formal tasks, data-diversity is not particularly important as long as the dataset requires a useful mechanism.
       \item A meta-training dataset is useful only if the neural-architecture is able to acquire the relevant mechanisms.
    \end{enumerate}
  \end{minipage}\\
   \bottomrule
  \end{tabular}
  \caption{\textbf{Predictions of the simplicity-bias view and the mechanistic view}. We find evidence in support of the mechanistic-complexity view (predictions (2a) and (2b)) and against the simplicity-bias view (predictions (1a) and (1b)).
    \label{tab:pred}
  } 
  \end{center}
\end{table*}

\section{Meta-Learning a simplicity prior?}
The received wisdom today is that the best kinds of generalizations are those that are implied by the Kolmogorov complexity of the data \citep{solomonoff_formal_1964,kolmogorov_tables_1963,chater_search_1999,chater_simplicity_2003}.
The idea is roughly that the shortest program which can generate your data will have the right kinds of generalizations.
While Kolmogorov complexity itself is non-computable, different approximations of it have been used before for neural-networks such as Minimum Description Length \citep{lan_minimum_2022}.
Unfortunately, these approximations tend themselves to be non-differentiable, leading them to be unsuitable for gradient-descent and large-scale connectionist approaches.

Within cognitive science, Bayesian priors for simplicity are deeply connected to a popular view of human cognitive processes, Rational Analysis \citep{anderson_is_1991}.
On this view, psychological functions are seen as adaptive solutions to challenges faced by organisms in their environment.
Under Rational Analysis, the idea that minds are adapted to their environment is identified with the claim that mental processes are \emph{rational} (in other words, for Rational Analysis, they follow a Bayesian prior).
This crucially entails that mental processes cannot just be fitting representations to their environment, they must also take into account the prior probabilities of said representations, including in particular a measure of the simplicity of those representations.
\citeposs{yang_one_2022} model sits squarely within this tradition.

Consequently, the Bayesian view of cognition is suspicious of approaches to learning which lack a rational analysis in this sense, as mainstream neural-network approaches do.
This suspicion is warranted on independent cognitive grounds: mainstream techniques are not only not Bayesian, in that they have no rationally-analyzable nor easily interpretable priors, they also require massive amounts of data to achieve the impressive behavioral results we've come to expect of them.
This relentless hunger for data is impossible to square with how children learn.
For example, in the case of natural language, the data is five to six orders of magnitude lower for children than the amount of data required to train modern language models \citep{hu_findings_2024}.
In sum, the idea that human learners come into the world equipped with a sophisticated prior probability reflecting, among other things, a bias for simplicity is often taken to provide both a straightforward explanation of why minds are as they are from an adaptationist perspective, and a promising avenue to address the age-old puzzle of how humans manage to learn so much from so little.

But there is a catch.
Bayesian update as recommended by rational analysis in this sense is intractable (and perhaps impossible in the case of infinite hypothesis spaces).
Even tractable approximations of Bayesian inference can only be done usefully within extremely restricted, orderly domains.
Despite \citeposs{yang_one_2022} great success on simple formal languages, their model's reliance on intractable, non-differentiable search makes it impossible to scale up to the full complexity of natural language.

The field is left in a frustrated position.
On the one hand, Bayesian inference offers a promising way of reproducing and explaining in one fell swoop both \emph{why} minds are the way they are and \emph{how} minds manage to draw sweeping generalizations on the basis of extremely limited data.
On the other hand, this approach is currently completely impracticable for the kinds of data that human minds \emph{do indeed} handle, while altogether non-Bayesian machine-learning approaches achieve impressive behavioral accuracy, crucially with serious issues in out-of-distribution generalization.

To resolve this tension, one could hope that a neural network could still be forced to follow some kind of simplicity prior.
This could be done by architectural constraints, but meta-learning provides a tantalising possibility: what if the simplicity prior could be meta-learned \citep{binz_meta-learned_2024}?
Perhaps by meta-training on a dataset which follows that simplicity prior, the model could learn to follow it.
This is how \citet{mccoy_modeling_2023} describe what their model is doing. 
In particular, the view claims the successes of meta-learning are born out of this bias from simplicity that is learnt from the training datasets' own bias for simplicity.

\subsection{Meta-learning a prior or the prior?}
One justification for the view that meta-learning is distilling a prior comes from a theoretical result by \citet{grant_recasting_2018}.
The authors shows that MAML can be interpreted as a hierarchical Bayesian model where MAML estimates a prior which is parameterized by the weights of the neural network being used.
A crucial detail about this theoretical result is that it is not guaranteed that this prior will approximate the data distribution nor that this prior will be interpretable. 
Rather, the learnt prior depends entirely on the specific neural architecture used and simply reflects the ideal initialisation for the specific neural network given the data provided, rather than an approximation of the underlying task distribution.
In other words, \citet{grant_recasting_2018} have shown that \emph{a} prior is learnt, but not necessarily the desired prior. 

Within Rational Analysis, this could be useful for some kinds of tasks, such as categorization. 
When recognising animals in images, for example, we could learn a prior about underlying frequencies of features of animals.
MAML could model this and produce an uninterpretable prior over image recognition models (with the important caveat that such priors depend on the neural architecture). 

Our focus, however, is on the thornier issue of generalization. 
While meta-learning could be interpreted as a way of imbuing various kinds of priors \citep{binz_meta-learned_2024,grant_recasting_2018}, if one is interested in generalization or higher-order cognitive processes, the received view is that the best prior would be for simplicity~\citep{binz_meta-learned_2024,chater_search_1999}.
Conversely, other priors (such as a prior for extra complexity) should lead the models to worse performance.
As such, it is crucial that the meta-learned model is acquiring \emph{the} prior that the model is exposed to, rather than just any old prior if we want to attribute the success of MAML models to their distillation of a specific prior (in this case, a simplicity prior).

\section{Meta-learning neural mechanisms!}
We have a different interpretation, whereby the model is not learning to follow a prior for simplicity, but is instead learning basic neural mechanisms which are useful on later tasks. 
Neural mechanisms are basic circuits; an example could be counters which have been found in LSTMs \citep{weiss_practical_2018,suzgun_lstm_2019}. 
These basic neural mechanisms can be learned from simple datasets provided the data require using these computational tools.
Since the meta-learned weights already contain useful mechanisms, they are much more easily found in parameter space when training than in randomly-initialized networks, enabling learning on smaller datasets with better generalizations.

Rather than organizing a meta-learning dataset around a \textbf{simplicity bias} as recommended by the Bayesian view of human cognition, we suggest organising it around \textbf{mechanistic complexity}.
We define mechanistic complexity as the amount of expressive power required to do a task as is done in formal language theory or automata theory.
For example, the formal language $ab, abab, ababab, \ldots$ can be considered simpler than the formal language, $ab, aabb, aaabbb, \ldots$ because the former can be done with a finite-state machine, while the latter requires a pushdown automaton.
From an information complexity point of view, these two languages may be roughly equivalent in their complexity, but in terms of mechanistic complexity, they are drastically different.

While neural architectures vary in terms of their expressive power, sometimes the mechanisms required to maximally utilise this expressive power can be hard to find in the parameter space.
For example, LSTMs are capable of counting \citep{gers_lstm_2001}, but there is no guarantee that a given LSTM will learn a counting mechanism if the provided data aren't sufficiently rich (and gradient-desscent made lead to suboptimal mechanisms \citep{lan_bridging_2024}).
We hypothesise that meta-learning on tasks with greater mechanistic complexity will lead to better models than meta-learning on less mechanistically complex tasks.
This would be because meta-learning encourages learning mechanisms (e.g.\ for counting) which then become easily accessible when training after meta-training.

\section{Methods}

We created two kinds of meta-learning datasets organised according to each principle.
For the simplicity-bias view, we created a collection of datasets varying in their information-complexity ranging from a simplicity prior to a complexity-prior.
For the mechanistic-complexity view, we used different formal languages as datasets and categorised them according to the Chomsky-hierarchy.

All our models were 2-layer 1024-dimensional LSTMs, following the parameters of \citet{mccoy_modeling_2023}.
Full hyperparameters can be found in Section~\ref{hyper} of the appendix.

\subsection{Simplicity bias with informationally-complex datasets}

To test whether simplicity is important for determining the quality of a meta-learning dataset, we created a dataset of 5000 formal languages.
These languages were generated using the Minimalist Grammar formalism \citep{stabler_derivational_1997} and consisted of simple formal languages à la \citet{yang_one_2022}.
Crucially, these languages were uniformly distributed in terms of their model description length (MDL; \citet{rissanen_modeling_1978}) from 0 to 100.
MDL was calculated as a function of the number of bits required to encode all features in the underlying grammar representation as in \citet{ermolaeva_learning_2021} (full details in Figures~\ref{fig:mdl} and \ref{fig:grammars}).

Under the assumption that a preference for simplicity drives generalisation, we should expect Bayesian models with a preference for complexity to perform \emph{worse}.
For example, \citet{yang_one_2022} point out that a prior for simplicity could drive the learning of $a^n$ rather than $a^{\{1,2,3\}}$ given the dataset $\left\{a,\ ,aa,\ aaa\right\}$.
This is because memorisation ($a^{\{1,2,3\}}$) is more complex than the generalised rule ($a^n$). 
A Bayesian model with a preference for some complexity would not make the right generalisation in this case and so they would perform worse than a model with a bias for simplicity.
Therefore, if the meta-trained model is really acquiring the prior, the models trained on a complexity prior should perform much worse than the ones trained on a simplicity prior.

\begin{figure*}[htb]

  \begin{subfigure}{\textwidth}
  \begin{center}
    \includegraphics[width=\textwidth]{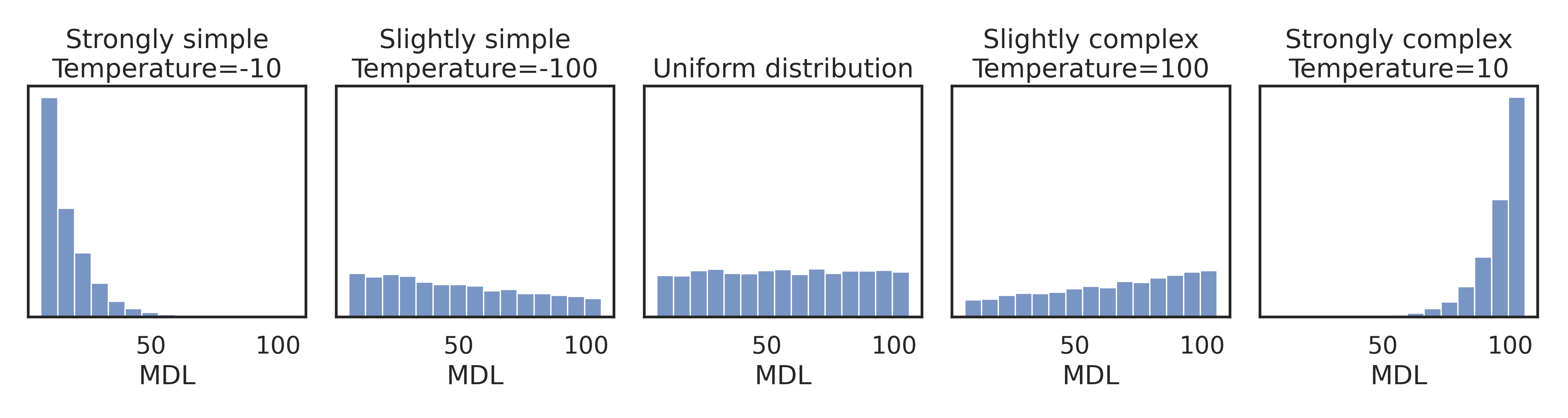}
  \end{center}
  \caption{The distribution of grammar complexity with different priors\label{fig:complexityhist}.
  }

  \end{subfigure}

  \begin{subfigure}{\textwidth}

    \begin{center}
\begin{tabular}{p{4cm}lcc}
\toprule
  Language & Examples & Family & Chomsky hierarchy\\ 
\midrule
  \texttt{a\^{}n} & a, aa, aaa & Growing & Regular \\
  \texttt{a\^{}nb\^{}n} & ab, aabb, aaabbb & Growing & Context-Free\\
  \texttt{a\^{}nb\^{}nc\^{}n} & abc, aabbcc, aaabbbccc & Growing & Context-Sensitive\\
  \midrule
  \texttt{kleene} & a, b, cba & Copy & Regular \\
  \texttt{wwR} & a$\vert$a, b$\vert$b, abab$\vert$baba& Copy & Context-Free\\
  \texttt{ww} & abc$\vert$abc, abab$\vert$abab, abac$\vert$abac  & Copy & Context-Sensitive\\
  \midrule
  \texttt{()\^{}n} & (), ()(), \{\}\{\}\{\} & Dyck & Regular \\
  \texttt{dyck} & (), (\{\}), ()() & Dyck & Context-Free\\
  \texttt{cross-dependency dyck} & (), \{(\})(), \{()\} & Dyck & Context-Sensitive\\
\bottomrule
\end{tabular}
\end{center}

\caption{All formal languages evaluated in this work. \label{fig:formallangs} }
  \end{subfigure}

  \caption{\textbf{Methodological Approach:} (a) \textbf{Simplicity-Bias View:}  We created several datasets where we manipulated the Minimum Description Length (MDL) of the languages. 
    The probability of sampling a grammar was calculated by taking the softmax of all the MDL scores of all 5000 grammars and the simplicity/complexity preference was set by manipulating the temperature. 
    The MDL calculation and example grammars are in the appendix.
  (b) \textbf{Mechanistic-Complexity View:} We created several datasets based on the Chomsky hierarchy to manipulate the mechanistic complexity required to learn the language. Languages were grouped into ``families'' based on superficial similarities and shared alphabets.\label{fig:details}}
\end{figure*}

Each meta-learning model was trained by sampling formal languages from this dataset.
Crucially, we could manipulate the frequency of simple or complex languages seen in training by sampling according to the MDL score of each grammar (see Figure~\ref{fig:details} for details).
This allowed us to test whether models performed better when they were trained to have a simplicity bias or a complexity bias.\footnote{Since complexity is right-unbounded, this is really more of a bias for ``medium'' complexity.}

\subsection{Mechanistically-complex datasets}
The mechanistic-complexity datasets were much more restricted than the information-complexity datasets.
Each meta-learning training regimen consisted of a \textbf{single language} chosen according to its mechanistic complexity.
If MAML really is distilling a prior into the model, then meta-learning a single language should lead to catastrophic performance as all the probability mass of the prior will be on that language.
Conversely, if MAML leads to the acquisition of neural mechanisms, then a single language could provide a useful meta-learning dataset, provided it encourages the learning of relevant mechanisms.

The datasets we constructed to test this predict consist of 9 formal languages.
These formal languages are described in Figure~\ref{fig:formallangs}.
The 9 languages were selected to vary in terms of their position in the Chomsky hierarchy, a classic measure of computational expressiveness, with three regular languages, three context-free languages and three context-sensitive languages.\footnote{Technically, all regular languages are also context-free and context-sensitive and likewise all context-free languages are also context-sensitive. For brevity, we describe a language by the least expressive position in the hierarchy that can express the language.}
The languages were also organised into ``families'', informal groupings to capture underlying similarities and shared vocabularies.

Languages were also chosen so that their strings had natural ``lengths'', corresponding roughly to the number of steps necessary to generate the string.
For example, the string $aabb$ in $a^nb^n$ has a length of two, while the string $((()()))$ in the Dyck language has a length of four.
Full definition of length for each language can be found in the appendix.

During meta-training, each training string is sampled as follows.
First, a length is sampled uniformly from 1 to 10, then from each length, a random string is sampled uniformly for that length.
The meta-learning test strings (used at the end of the inner loop) are sampled in the same manner but with lengths from 11 to 20.
This was done to help length generalization and as an easy way to ensure the inner-loop test and training set do not overlap.

Finally, during meta-learning, the symbols were randomly assigned to different vocabulary indices for each inner loop (excluding \textsc{[Start]}, \textsc{[Stop]} and \textsc{[Pad]} tokens which had consistent indices).
So, while $a$ might correspond to an index of $3$ in one batch, it could correspond to $5$ in another batch.
This was done to encourage generalisation, to ensure that each word-embedding was initialised and to allow the model to learn languages with larger vocabularies than their meta-learning language.
By shuffling vocabulary indices from batch to batch, each batch is like a different task albeit on the same basic language, showing how we are meta-training rather than pretraining on a single language.

\subsection{Training after meta-training}
After meta-training, all models were trained on the same data.
Rather than evaluate on the 56 formal languages used by \citet{yang_one_2022}, we used the same target languages as in Figure~\ref{fig:formallangs} because of their various formal properties.
This allowed us to define a more precise evaluation scheme, described in the following section.

To train models, we followed the exact same scheme as
\citet{mccoy_modeling_2023}, where the meta-trained model is trained for $n$ epochs using SGD and then $m$ epochs using Adam.
The full details are available in Table~\ref{tab:train} in the appendix.

All models were trained with the same sampled data from each formal language where $n$ strings were sampled by first uniformly sampling a length, and then uniformly sampling a string for that length for a formal language.

\subsection{Evaluation}

\paragraph{Yang and Piantadosi's F1}
To evaluate the different models, we used a slightly different technique than \citet{yang_one_2022} or \citet{mccoy_modeling_2023}.
\citet{yang_one_2022} introduced a modified F1 score as a metric to track the acquisition of a formal language.
Since many formal languages are defined by infinite sets, it is tricky to determine whether a model has correctly acquired a language without a formal parser for each language.
Rather than formally proving a language has been acquired, Y\&P simply took the 25 most probable strings from both the model and target language and used them to create a modified F1 score.
`Precision' was defined as the proportion of the 25 most likely strings in the model that were members of the target language (defined as a very large corpus of sampled strings).
`Recall' was the proportion of the 25 most likely strings in the target language that the model had sampled after sampling many strings from the model.

This metric is not sensitive to length and cannot allow us to tell if a model has generalised (since sufficiently long strings will simply not be in the top 25).
Furthermore, it presupposes a probabilistic ordering of strings.
While it seems natural to assume that $a$ is more likely than $aaaaa$ in $a^n$, this needn't be the case, and for other grammars it is less straightforward to decide on an appropriate distribution.
Technically, formal languages are simply sets of strings so imposing an ordering on them amounts to making extra assumptions which may not be warranted.

\paragraph{Continuations}
Our solution is to use a different metric which looks at \textbf{the degree to which the model correctly \emph{continues} strings.}
Given a formal language, we can look at each string, token by token, and see what are valid next tokens in the formal language.
For example, given the language $a^nb^n$, if the string so far is $aa$, the next valid token could be $a$ or $b$.
However, if the string is $aab$, the only valid token is $b$.
A similar strategy has been used to great effect in analysing language models' acquisition of grammatical rules in natural language \citep{linzen_assessing_2016}.

\begin{table}
  \begin{center}
  \begin{tabular}{lcr}
    \toprule
    String   & Length & Valid continuations        \\
    \midrule
    (        & 1      & $(,\ \{,\ )$               \\
    ()       & 1      & $(,\ \{,\ \textsc{[Stop]}$ \\
    ()\{     & 2      & $(,\ \{,\ \}$              \\
    ()\{(    & 3      & $(,\ \{,\ )$               \\
    ()\{()   & 3      & $(,\ \{, \}$               \\
    ()\{()\} & 3      & $(,\ \{, \textsc{[Stop]}$  \\
    \bottomrule
  \end{tabular}
  \end{center}
  \caption{All valid continuations generated from the string $()\{()\}$ in the Dyck language. \label{tab:cont}}
\end{table}

Our evaluation corpus was built by sampling 10 strings for each of our target languages from each length from 1 to 40.
For every generated string, we included all possible continuations for each token in the string (see Table~\ref{tab:cont} for an example).

For each continuation, we can then formalize various metrics to determine a model's acquisition of a language.
We designed our metrics to be analogous to \citeposs{yang_one_2022} variant of the F1 score.

For precision, the most natural definition is the likelihood of a model choosing a valid token.
Let $s$ be a string and $\textsc{Val}(s)$ be the set  of valid continuations given $s$.
Finally, let $P(x \mid s)$ be the probability the model assigns to token $x$ given string $s$.
\begin{align*}
  P_\text{Val}(s) &= \sum_{x\in\textsc{Val}(s)} P(x \mid s)
\end{align*}

This will not be a sufficient metric since a degenerate model can still have perfect precision (for example, a model which always predicts ``('' could have perfect precision in Table~\ref{tab:cont}).

For an analogy to recall, we defined a metric we call ``better-than.''
It is defined simply as the proportion of valid moves which have a higher probability than all invalid moves combined.
We considered alternative metrics as well; these can be found in the appendix.

\begin{align*}
  \text{BT}(s) &= \frac{\sum\limits_{x\in\textsc{Val}(s)} \left[ P(x \mid s) > \sum\limits_{c\not\in\textsc{Val}(s)} P(c \mid s) \right]}{\left\lvert\textsc{Val}(s)\right\rvert}
\end{align*}

Finally, we define our variant of F1 as the harmonic mean of these two measures:
\begin{align*}
  F1(s) &= \frac{2 P_\text{Val}(s)\cdot \text{BT}(s)}{P_\text{Val}(s) +  \textsc{BT}(s)}
\end{align*}
Note that this term is defined for each continuation, rather than for an entire language.

\section{Results}

\begin{figure*}[htb]
  \begin{center}
  \includegraphics[width=\textwidth]{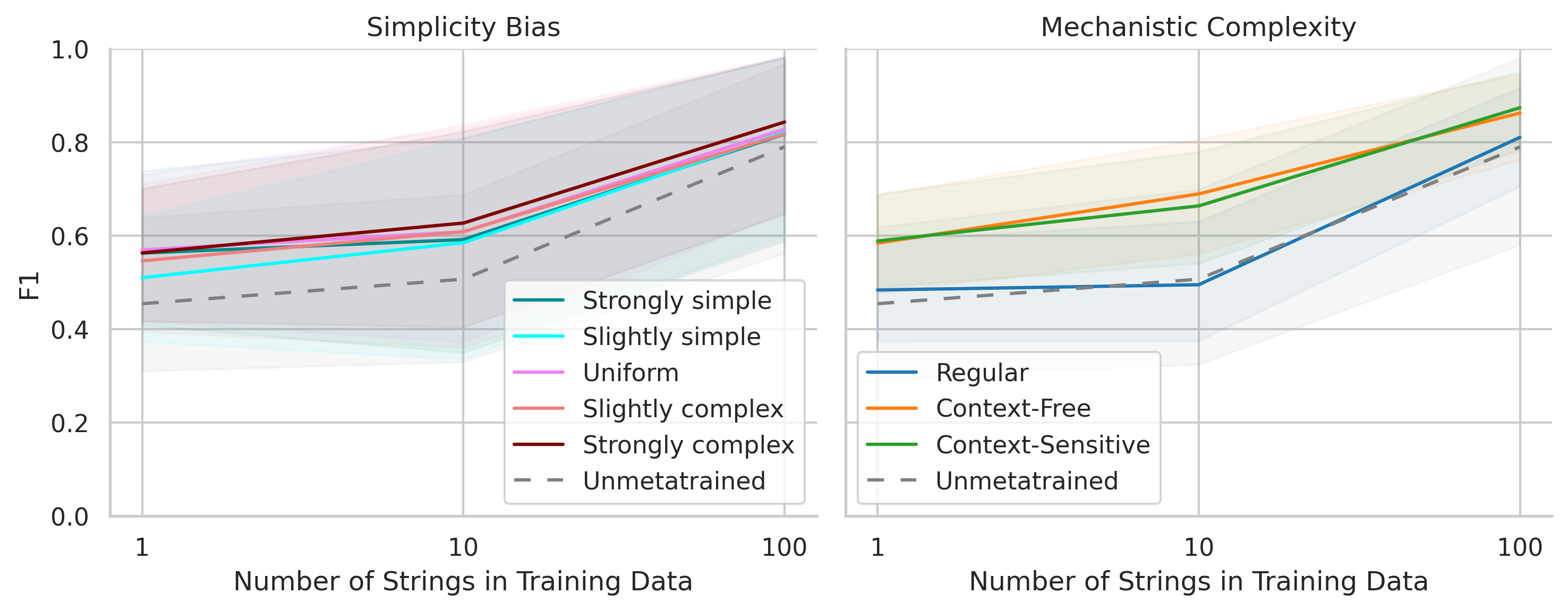}
  \end{center}
  \caption{\textbf{The mechanistic view but not the simplicity-bias view accounts for meta-learning effects}.  Mean F1 for each target language for continuations with length $\leq 10$ (averaged over three meta-learning runs).
    All models were trained on the same data sampled from each target language after meta-training.
    Mechanistic complexity excludes cases where the target language is the same as the meta-training language.
    Variant figures with different metrics are available in the appendix.
    \textbf{Preferring simple or complex languages shows no difference in model performance whereas meta-training on more mechanistically-complex languages (context-sensitive/context-free languages) is more helpful than meta-training on less mechanistically-complex ones (regular languages).}
    \label{fig:f1_time}
  }
\end{figure*}

Overall, we found that models meta-trained on context-free and context-sensitive languages outperformed models meta-trained on regular languages (see Figure~\ref{fig:f1_time}).
Against Bayesian intuitions, we found that models meta-trained with a simplicity-bias did \emph{not} outperform models meta-trained with a complexity-bias, or for that matter models meta-trained on a single context-free or context-sensitive language.
This is despite the fact that the informationally-complex datasets consist of 5000 languages whereas the mechanistically-complex datasets had a single language.
This result indicates that, for this meta-training scenario, more diverse data does not necessarily entail more performant models.

\subsection{Simplicity bias does not predict performance}

We found that there was no real distinction between meta-learning models that were trained on a dataset with a simplicity-bias or with a complexity-bias (Figure~\ref{fig:f1_time}).
The meta-trained models do perform much better than unmetatrained models, but there do not appear to be any significant differences on whether the model is meta-trained on simple or complex data.
This directly contradicts the predictions of the simplicity-bias view (Table~\ref{tab:pred}).

\subsection{Mechanistic-complexity does predict performance}

\begin{figure*}[htb]

  \includegraphics[width=\textwidth]{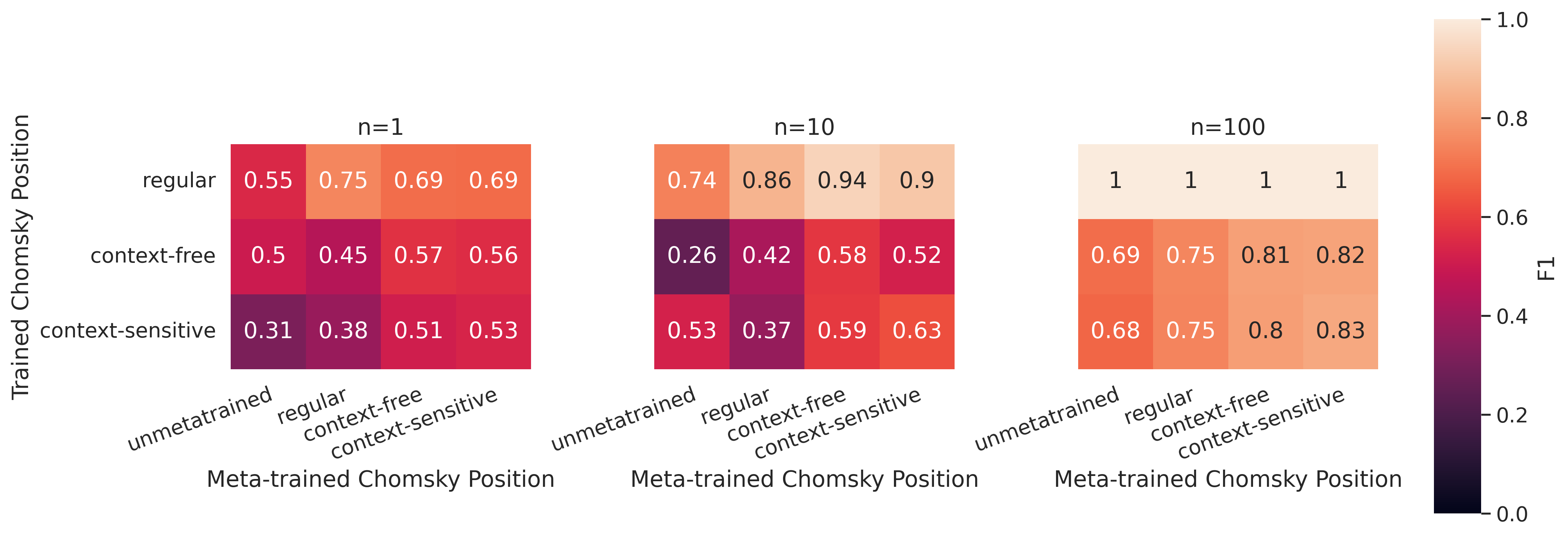}
  \caption{
    \textbf{Meta-learning on languages high in the Chomsky hierarchy helps learning those lower in the Chomsky hierarchy but not vice-versa.}
    Here we show the generalisation of different models across levels of the Chomsky hierarchy.
    Of particular interest, with ten training strings, models meta-trained on context-free and context-sensitive languages are better at learning regular grammars compared to models meta-trained on regular grammars. 
    Conversely, meta-training on regular grammars do not seem to provide very much improvement over unmetatrained models.
    Each value is the mean F1 across target languages for continuations with length $\leq 10$.
    \label{fig:confusion}
  }
\end{figure*}

Contrary to the simplicity-bias view, we found a clear differentiation among models meta-trained with different mechanistic complexity.
The models meta-trained on regular languages performed worse than those meta-trained on context-free or context-sensitive languages.
In fact, the regular models were not consistently better than unmetatrained models (Figures~\ref{fig:f1_time} and \ref{fig:confusion}).
Figure~\ref{fig:confusion} shows how meta-training on languages higher in the Chomsky-hierarchy helps learning languages lower in the Chomsky-hierarchy but the converse is not true. 
This goes with the prediction (2a; Table~\ref{tab:pred}) as going up each level in the Chomsky-hierarchy necessarily requires a more expressive mechanisms.

The success of meta-training on a single-language vindicates prediction (2a) of the mechanistic-complexity view.
If meta-learning inculcates useful neural mechanisms, then a single, sufficiently complex language (such as $a^nb^nc^n$) should suffice to learn it.
Conversely, meta-learning mechanistically-\emph{simple} languages does not seem to help, as regular languages do not bestow any useful mechanisms.

\begin{figure}[htb]
  \begin{center}
  \includegraphics[width=0.9\columnwidth]{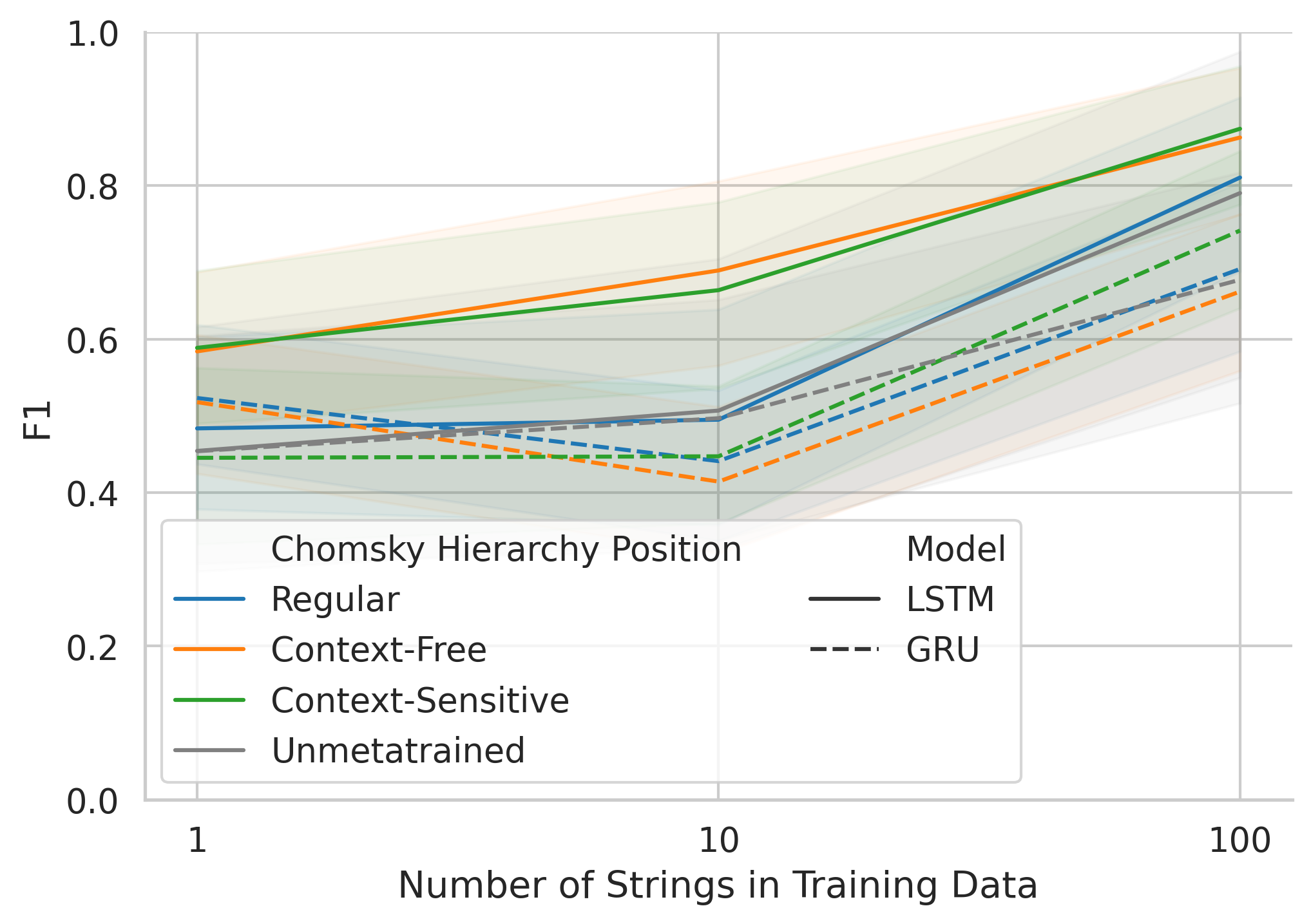}
  \end{center}
\caption{\textbf{GRUs do not show the same meta-learning effect as LSTMs}. GRUs do not show improvement from meta-training. As GRUs cannot express counters in constrast to LSTMs \citep{weiss_practical_2018}, they might not learn useful neural mechanisms during meta-training}\label{fig:gru}
\end{figure}

In particular, we hypothesise that the LSTMs are learning mechanisms for counters, as this has been previously shown~\citep{weiss_practical_2018}, and this is sufficient for $a^nb^nc^n$. 
To test this, we looked at GRU models which cannot acquire counters \citep{weiss_practical_2018}, and we found they had no benefit from meta-training on the same datasets as the LSTMs (Figure~\ref{fig:gru}).
This follows the prediction (2b) of the mechanistic-complexity view (table~\ref{tab:pred}) where we expect neural architectures that cannot acquire the relevant neural mechanisms to not benefit from meta-training on that task.

\section{Discussion}

Our results indicate that meta-trained models don't directly mimic the statistical properties of their meta-training dataset.
Whether meta-trained on informationally-simple or informationally-complex datasets, they perform equally well on later training tasks.
This casts doubt on the idea that meta-trained models succeed due to the acquisition of a simplicity-bias and implies that MAML does not bestow a specific prior on a neural network.

Conversely, meta-training on mechanistic-complex languages \emph{does} seem to be vital in encouraging models to acquire useful neural-mechanisms.
Despite being a single language dataset, the context-free and context-sensitive models were better than the simplicity-bias models which were trained on 5000 languages.
We hypothesise this is the result of the context-sensitive/context-free languages providing a computationally richer base than informationally-complex languages (which can be complicated yet be a regular or even finite languages) prompting the meta-learned models to learn basic neural mechanisms.

While we found a clear differentiation between regular languages and context-free or context-sensitive languages, we did not between context-sensitive and context-free languages.
This likely has to do with the relative imprecision of the Chomsky hierarchy; there are many important differences between languages that aren't captured by the Chomsky hierarchy.
For example, $a^nb^n$ and $a^nb^nc^n$, require only tracking the number of items on the stack (and whether we have moved from $b$'s to $c$'s for $a^nb^nc^n$), rather than actually keeping track of the specific items and the order they occur in.

Conversely, the copy languages require perfectly recording each symbol on the stack.
The models failed to completely learn the copy language, despite succeeding on $a^nb^nc^n$.
The relative difficulties of these languages corresponds with what we know about the capacities of LSTMs.
LSTMs can acquire counters to keep track of cardinalities \citep{weiss_practical_2018}, whereas learning a fully-fledged stack may prove more difficult or practically impossible \citep{deletang_neural_2023}.

As such, the Chomsky-hierarchy may not be the best way to characterise the cognitive ``primitives'' (i.e.\ basic neural mechanisms enabling generalization) that are most useful for a neural architecture.
Indeed, GRUs show no improvement from meta-training at all, since they cannot acquire the counting mechanisms that LSTMs are able to exploit after learning $a^nb^nc^n$. 
This follows previously theoretical and empirical work showing that, while LSTMs can acquire counting languages \citep{fischer_counter_1968}, GRUs cannot \citep{weiss_practical_2018}. 

Future work could examine what alternative theoretical hierarchies like the Chomsky-hierarchy best correspond to the neural mechanisms that are useful in an architecture.
For example, many recent works have looked at the expressivity of Transformer models \citep{hao_formal_2022,strobl_what_2024,yang_counting_2024,yang_masked_2024,zhou_what_2024}, allowing one to define a ``circuit-complexity hierarchy.''
Outside of meta-learning, previous works have found that \emph{pre}-training on formal languages or other sources of structure (e.g.\ music) can help later learning of natural language~\citep{papadimitriou_learning_2020,papadimitriou_injecting_2023}; this may be due to similar reasons.

Given a better understanding of the mechanisms that can be learnt by different neural architectures, meta-learning could be used to initialise models with initial parameters that make useful generalisations more accessible in the parameter-space.
Just as languages like $a^nb^nc^n$ encourage LSTMs to learn a counting mechanism, different meta-learning tasks might, for example, encourage Transformers to learn how to use their positional encodings ahead of actual training.
Work like RASP \citep{weiss_thinking_2021} which define different possible circuits in neural architecture can provide an interesting basis for seeing what kinds of neural mechanisms would be useful to meta-learn in a specific neural architecture.

\section*{Limitations}
This article addresses meta-learning in a fairly narrow domain, formal language learning, without looking at more complicated domains such as natural language.
While we argue that neural mechanisms explain our results, and counters have been previously found in LSTMs (and not in GRUs) \cite{weiss_practical_2018}, we do not explicitly look at hidden layers to find the presence of a neural counter. 
Future work could look to investigate different architectures beyond LSTMs and GRUs---Transformers being an obvious possibility, but more exotic architectures such as Stack-RNNs remain intriguing. 
It would also be of vital interest to develop computational hierarchies akin to the Chomsky hierarchy but defined in terms of specific neural architectures and the neural mechanisms that can be found in them.

\section*{Acknowledgements}
We thank Nur Lan and Emmanuel Chemla for their helpful comments and discussion as well as the members of Michael Goodale's thesis advisory committee: Tom McCoy, Louise McNally and Luigi Rizzi. 

This research was supported in part by Agence Nationale de la Recherche grant ANR-19-P3IA-0001 (PRAIRIE 3IA Institute; Natural Language Inference project, PI: Mascarenhas) as well as a grant from École Doctorale Frontières de l’Innovation en Recherche et Education—Programme Bettencourt.
Ecole Normale Supérieure-PSL’s Département d’Etudes Cognitives is supported by ANR grants, ANR-10-IDEX-0001-02 and FrontCog, ANR-17-EURE-0017.
This work was granted access to the HPC resources of IDRIS under the allocation {2024-AD011015605} made by GENCI.

\bibliography{meta-learning}

\appendix

\section{Meta-learning hyperparameters}\label{hyper}
All models were 2-layer LSTMs trained with a vocabulary size of 10 (including 3 tokens for \textsc{[Start]}, \textsc{[Stop]} and \textsc{[Pad]}) and an embedding and hidden dimension of 1024.
Models were trained with first-order MAML and were presented with 25 000 inner-loops with a batch-size of 1 and two gradient-accumulation steps.

The inner-loop had a learning rate of 1.0 using SGD, while the outer-loop was optimized with Adam and a learning rate of 0.0001.
Each inner-loop consisted of 200 training strings presented in 20 batches of 10.
The inner-loop test set consisted of 2 batches of 10 test strings.

Our hyperparameters were selected by a grid search over different parameters for hidden-size ($\{4,16,64,256, 1024\}$) and outer learning rate ($\{0.001, 0.0001, 0.00001\}$), and inner learning rate ($\{0.1, 1.0\}$).
The best parameters were chosen based on the perplexity of each meta-trained model on its own meta-trained language (e.g. a model meta-trained on $a^nb^nc^n$'s perplexity after being trained on $a^nb^nc^n$).
We could not evaluate all possible parameters due to VRAM limitations; in particular we were limited in batch-size and forced to use first-order MAML.

Our total compuational consumption was roughly 5000 hours of GPU time on Nvidia V100s. 
The code was developed using PyTorch 2.4~\citep{ansel_pytorch_2024}, PyTorch Lightning 2.1~\citep{falcon_pytorch_2019} as well as Higher~\citep{grefenstette_generalized_2019} for the MAML implementation.
This paper was written without the aid of AI assistants.

\begin{table}[htb]
  \begin{center}
  \begin{tabular}{ccc}
    \toprule
    $n$ strings & SGD Epochs & Adam Epochs \\
    \midrule
    1 & 5 & 1\\
    10 & 5 & 1\\
    100 & 10 & 5\\
    \bottomrule
  \end{tabular}
  \end{center}
  \caption{Hyperparameters for down-stream training with batch-size of 32, SGD learning rate of 1.0 and Adam learning rate of 0.0005. These hyperparameters were copied from \citep{mccoy_modeling_2023} \label{tab:train}}
\end{table}

\begin{figure*}[htb]
  \begin{center}
  \includegraphics[width=1\textwidth]{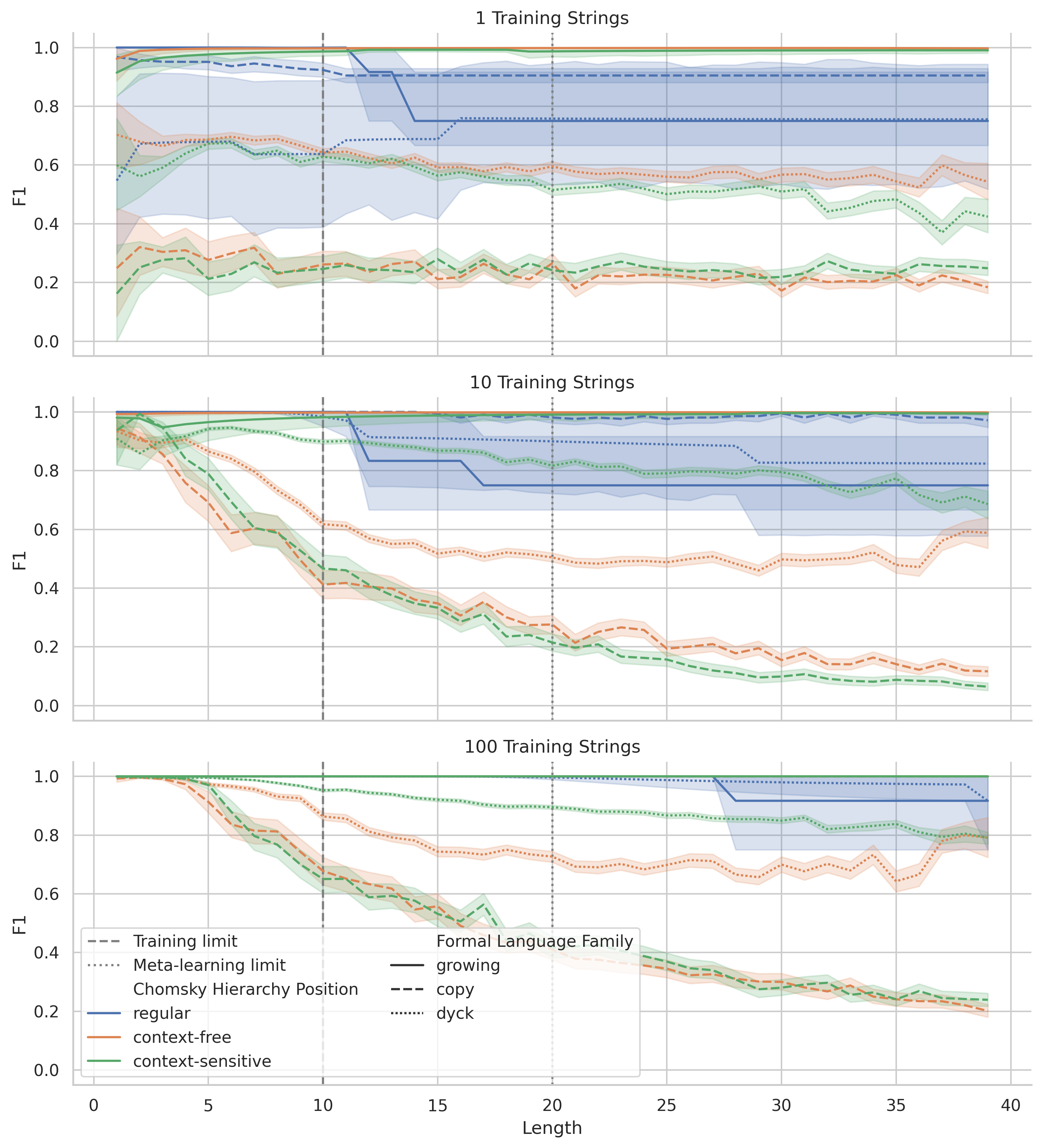}
  \end{center}
  \caption{Length generalization for each meta-trained model when trained on its own language. 
  Counting languages in family ``grow'' are learnt, while stack-based families ``dyck'' and ``copy'' suffer.
  \label{fig:length}
}
\end{figure*}

\begin{figure*}[htb]
  \begin{center}
  \begin{subfigure}[t]{0.5\textwidth}
    \includegraphics[width=\textwidth]{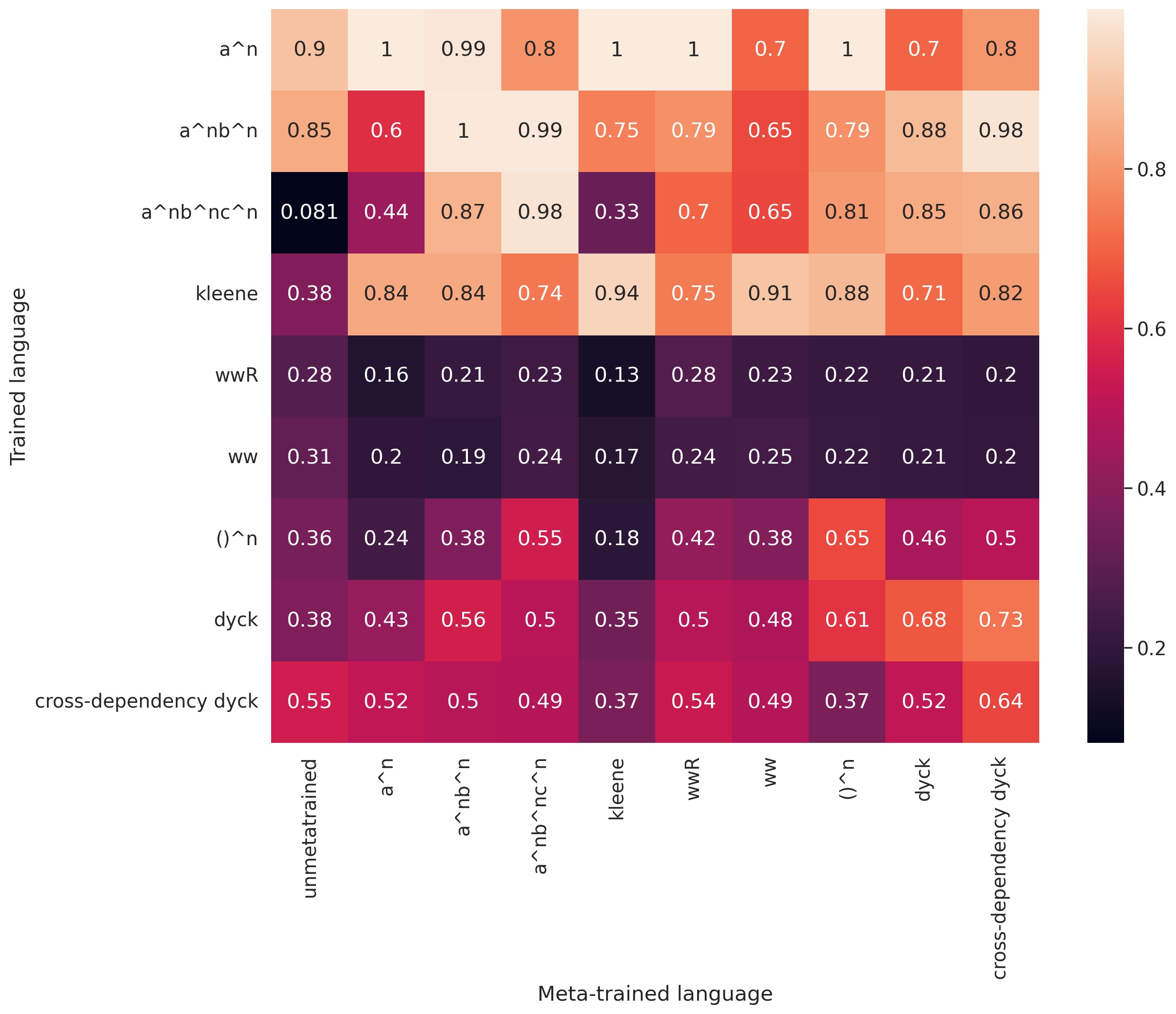}
    \caption{1 strings}
  \end{subfigure}
  \begin{subfigure}[t]{0.5\textwidth}
    \includegraphics[width=\textwidth]{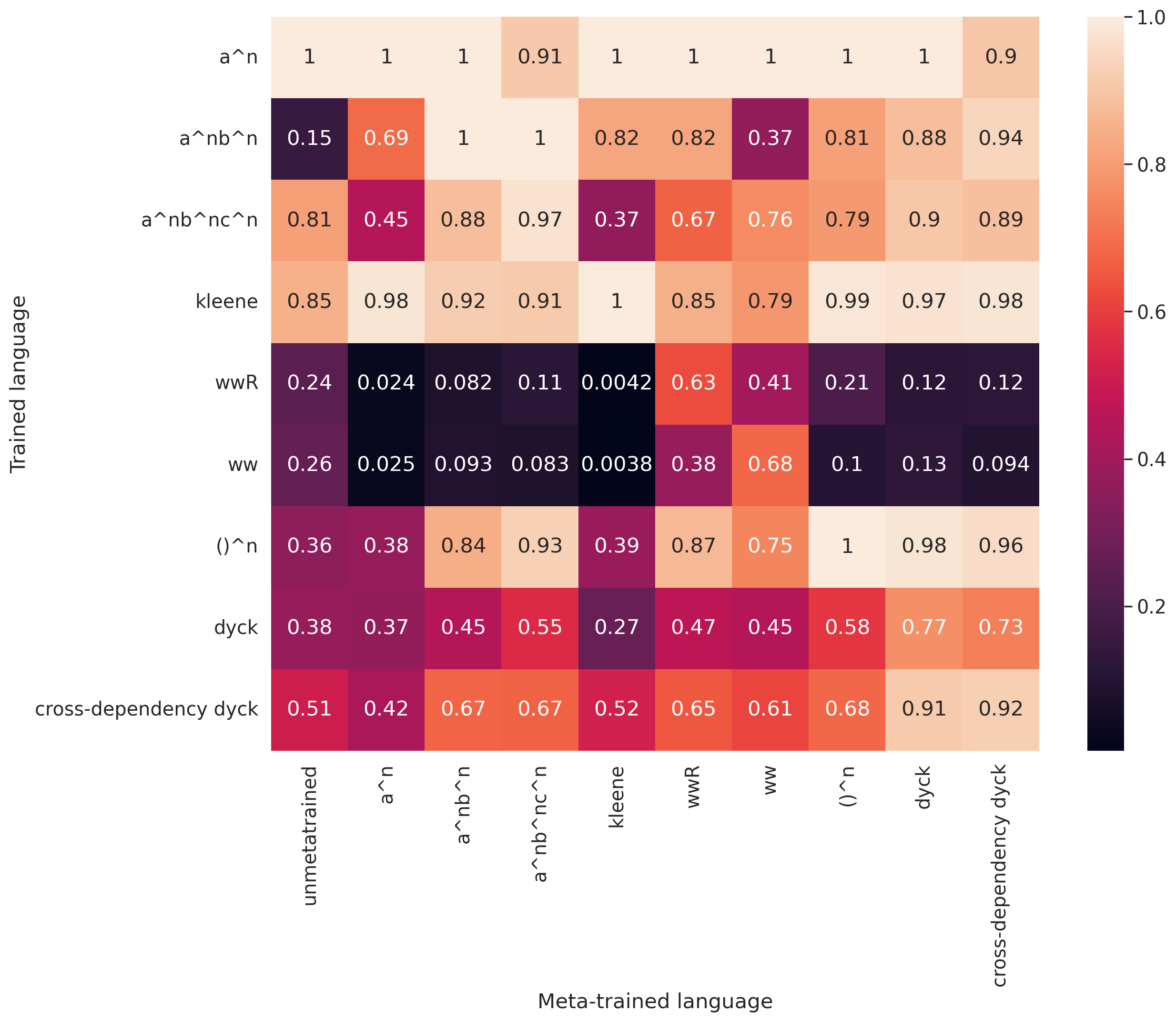}
    \caption{10 strings}
  \end{subfigure}

  \begin{subfigure}[t]{0.5\textwidth}
    \includegraphics[width=\textwidth]{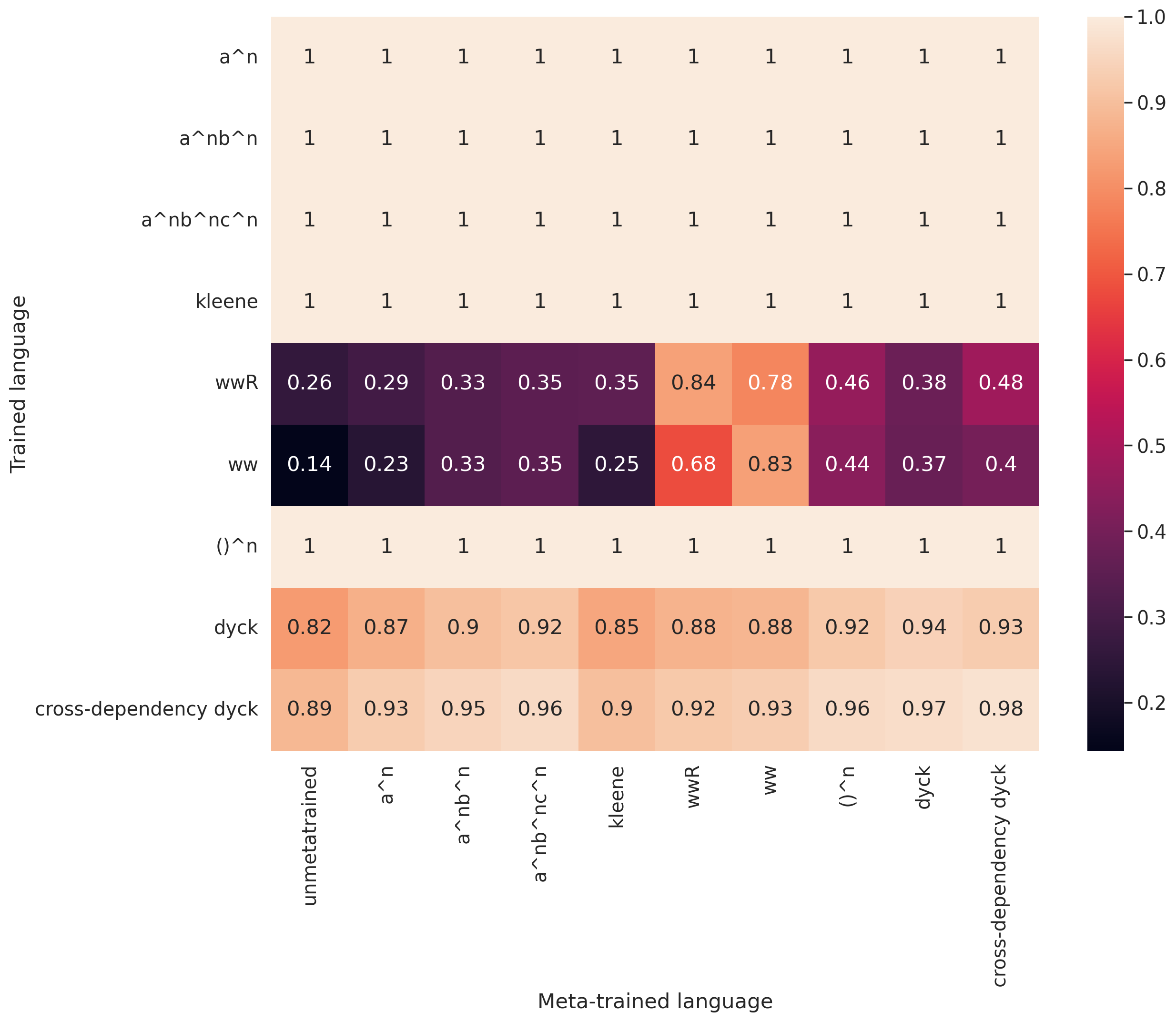}
    \caption{100 strings}
  \end{subfigure}
  \end{center}
  \caption{Mean F1 for all continuations with length $\leq 10$ for all formal languages}
\end{figure*}

\begin{table*}
  \begin{tabular}{cp{3cm}p{4.5cm}p{4.5cm}}
  \toprule
  \diagbox[width=3cm]{Family}{Chomsky} & Regular & Context-Free & Context-Sensitive \\
  \midrule
  \multirow{3}{*}{Growing} & \mC{$a^n$} & \mC{$a^nb^n$} & \mC{$a^nb^nc^n$} \\
  & Any number of $a$'s & Any number of $a$'s. followed by an equal number of $b$'s. & Any number of $a$'s followed by an equal number of $b$'s, then an equal number of $c$'s.\\
  & $a, aa, aaa,\ldots$ & $ab, aabb, aaabbb,\ldots$ & $abc, aabbcc, aaabbbccc,\ldots$ \\
  \midrule
  \multirow{3}{*}{Dyck} & \mC{$()^n$} & \mC{Dyck} & \mC{Cross-dependency Dyck}\\
                        & Any number of $()$ or $\{\}$ repeating. & Balanced parentheses or brackets & Dyck but treating brackets and parentheses independently.  \\
                        & $(), ()(), \{\} \{\}\{\}, \ldots$ & $(), (()), \{()\}, \ldots$ & $\{(\}), (\{)\}, ()(), \ldots$\\
                        \midrule
  \multirow{3}{*}{Copy} & \mC{Kleene} & \mC{Mirrored}& \mC{Copy} \\
                        & The Kleene star of $\{a,b,c\}$ & A string from Kleene followed by ``$\vert$'' and then the string reversed & A string from Kleene repeated twice and separated by ``$\vert$'' \\
                        & $a, bcd, cd, \ldots$ & $a\vert a, ab\vert ba, bcd\vert dcb$ &  $a\vert a, ab\vert ab, bcd\vert bcd$ \\
  \bottomrule
  \end{tabular}
  \caption{Detailed descriptions of each formal language used in this study.}
\end{table*}

\begin{table*}
  \begin{center}
  \begin{tabular}{cc}
    \toprule 
    Language & Length definition \\ 
\midrule
  \texttt{a\^{}n} & Number of $a$'s \\
  \texttt{a\^{}nb\^{}n} &  Number of $a$'s \\
  \texttt{a\^{}nb\^{}nc\^{}n} & Number of $a$'s \\
  \midrule
  \texttt{kleene} & Number of characters \\
  \texttt{wwR} & Number of characters prior to $\vert$ \\
  \texttt{ww} & Number of characters prior to $\vert$ \\
  \midrule
  \texttt{()\^{}n} & Number of parentheses/bracket pairs \\
  \texttt{dyck} & Number of parentheses/bracket pairs\\
  \texttt{cross-dependency dyck} &Number of parentheses/bracket pairs\\
  \bottomrule
  \end{tabular}
  \end{center}

  \caption{Definition of length for each language}
\end{table*}

\begin{figure*}[htb]

  \begin{align*}
    \text{MDL}(g) = \sum_{s::F \in g} \textsc{Not-}\epsilon(s)\log{p} + \left\lvert F\right\rvert\log{(5c(g))}
  \end{align*}

  \caption{MDL calculation for MGs adapted from \citet{ermolaeva_learning_2021} and \citet{katzir_cognitively_2014}. 
    $c(g)$ is a function which gives the number of categories in grammar, $g$.  
    $s$ and $F$ are the string and the features of a lexical entry respectively.
    $\textsc{Non-}\epsilon(s)$ is 1 if $s$ is not empty and 0 otherwise.
    $p$ is the number of possible lemmas for a grammar (7 in our case). 
    The 5 comes from the number of kinds of MG features (e.g. $\{$\texttt{=c, c=, c, -c, +c}$\}$ where \texttt{c} is some category). 
    \label{fig:mdl}
  }
\end{figure*}

\begin{figure*}[htb]
  \begin{center}
  \begin{subfigure}[t]{0.28\textwidth}
    \caption{MDL of approximately $7$}
    \begin{center}
    \textbf{Grammar}\\ 
      \lexeme{0}{0= 0}\\
      \lexeme{$\epsilon$}{0}

    \textbf{Top strings}\\

    \begin{small}
    $\{ \epsilon$, 0, 00, 000, 0000, 00000, 000000, 0000000, 0000000, 00000000, \ldots$\}$
  \end{small}
  \end{center}
  \end{subfigure}%
  \begin{subfigure}[t]{0.28\textwidth}
    \caption{MDL of approximately $52$}
    \begin{center}
    \textbf{Grammar}\\ 
    \begin{multicols}{2}
    \lexeme{4}{0}\\
  \lexeme{3}{1= 0}\\
  \lexeme{4}{1= 0}\\
  \lexeme{3}{1= 0}\\
  \lexeme{6}{1= 0}\\
  \lexeme{6}{=0 0}\\
  \lexeme{6}{=0 0}\\
  \lexeme{5}{1}\\
\lexeme{4}{1}\end{multicols}
    \textbf{Top strings}\\
    \begin{small}
    $\{$4,  46,  35,  34,  64,  44, 45,  65,  466,  346,  356,  646,  456,  656,  446, \ldots$\}$
  \end{small}
\end{center}
  \end{subfigure}%
  \begin{subfigure}[t]{0.34\textwidth}
    \caption{MDL of approximately $96$}
    \begin{center}
    \textbf{Grammar}\\ 
    \begin{multicols}{2}
      \lexeme{0}{3=~0=~+1~0}\\
      \lexeme{4}{0=~+2~+1~0}\\
      \lexeme{2}{3}\\
      \lexeme{6}{=4~=4~=0~0=~0~-2}\\
      \lexeme{1}{0=~0=~0~-2}\\
      \lexeme{3}{4~-2~-2}\\
      \lexeme{3}{4}\\
      \lexeme{6}{0~-1}
\end{multicols}
    \textbf{Top strings}\\ 
    \begin{small}
$\{$602,
616024,
66023364,
63366024,
616160244,
66160243364,
61660233644,
6660233643364,
61633660244,
63366160244, \ldots $\}$
\end{small}
\end{center}
  \end{subfigure}
  \end{center}

  \caption{A comparison of different grammars of different complexity used to train the simplicity-biased models.\label{fig:grammars} 
    In MGs \citep{stabler_derivational_1997}, grammars are defined as a set of lexical items. 
    Each lexical item has a string associated with it (to the left of the ``::'') as well as a sequence of features.
    These features determine which lexical items can merge with what. 
    For example, the simple grammar defines $a^n$ where the first lexical item merges other items of the category $0$ to its right.
    When merging, one can also move internal items with licensee features (e.g. \texttt{-c}) to licensor features (\texttt{+c}). 
    In the complex grammar, the ``6'' in front of all strings is the result of moving the lexical item \lexeme{6}{0 -1} to satisfy a \texttt{+1} feature.
  }
\end{figure*}

\begin{figure*}
\begin{align*}
  \text{BT}(s) = \frac{\sum\limits_{x\in\textsc{Val}(s)} \left[ P(x \mid s) > \sum\limits_{c\not\in\textsc{Val}(s)} P(c \mid s) \right]}{\left\lvert\textsc{Val}(s)\right\rvert}\\\text{Proportion of valid tokens which are more likely than all invalid tokens.}\\
  \textsc{BC}(s) = \sum_{x\in \textsc{Valid}(s)} \sqrt{\frac{P(x|s)}{\left\lvert \textsc{Valid}(s) \right\rvert }} \\\text{Bhattacharyya coefficient with uniform distribution over continuations.}\\\\
\textsc{Worst}(s) = {\left\lvert \textsc{Valid}(s) \right\rvert}\cdot {\text{Min}_{x\in\textsc{Valid}(s)} P(x|s)}\\\text{Least likely continuation scaled by number of continuations.}\\\\
\textsc{Possible}(s) = \frac{1}{\left\lvert\textsc{Valid}(s)\right\rvert}\sum_{x\in\textsc{Valid}(s)}\left\{\begin{array}{lr}
    1 &\text{ if } P(x|s) \geq \frac{\alpha}{\left\lvert \textsc{Valid}(s) \right\rvert} \\
    0 &\text{ if } P(x|s) < \frac{\alpha}{\left\lvert \textsc{Valid}(s) \right\rvert} \\
\end{array}\right\}
\\\text{Proportion of continuations with probability of at least $\alpha$ over the number of valid continuations.}
\end{align*}
\caption{\textbf{Variant metrics:}
In this paper, we evaluated our continuations using the ``better-than'' metric that we developed. 
However, there were alternative metrics we considered to replace recall in the F1 score.
These metrics showed the same qualitative result, except for the \textsc{Worst} metric. 
This is unsurprising as that metric is sensitive \emph{only} to least likely token, and so does not reflect partial successes well.
We reproduced Figure~\ref{fig:f1_time} with the different metrics in Figure~\ref{fig:variantplot}.
\label{fig:variant}}
\end{figure*}

\begin{figure*}
\centering
  \begin{subfigure}[t]{\textwidth}
    \centering
    \includegraphics[width=0.9\textwidth]{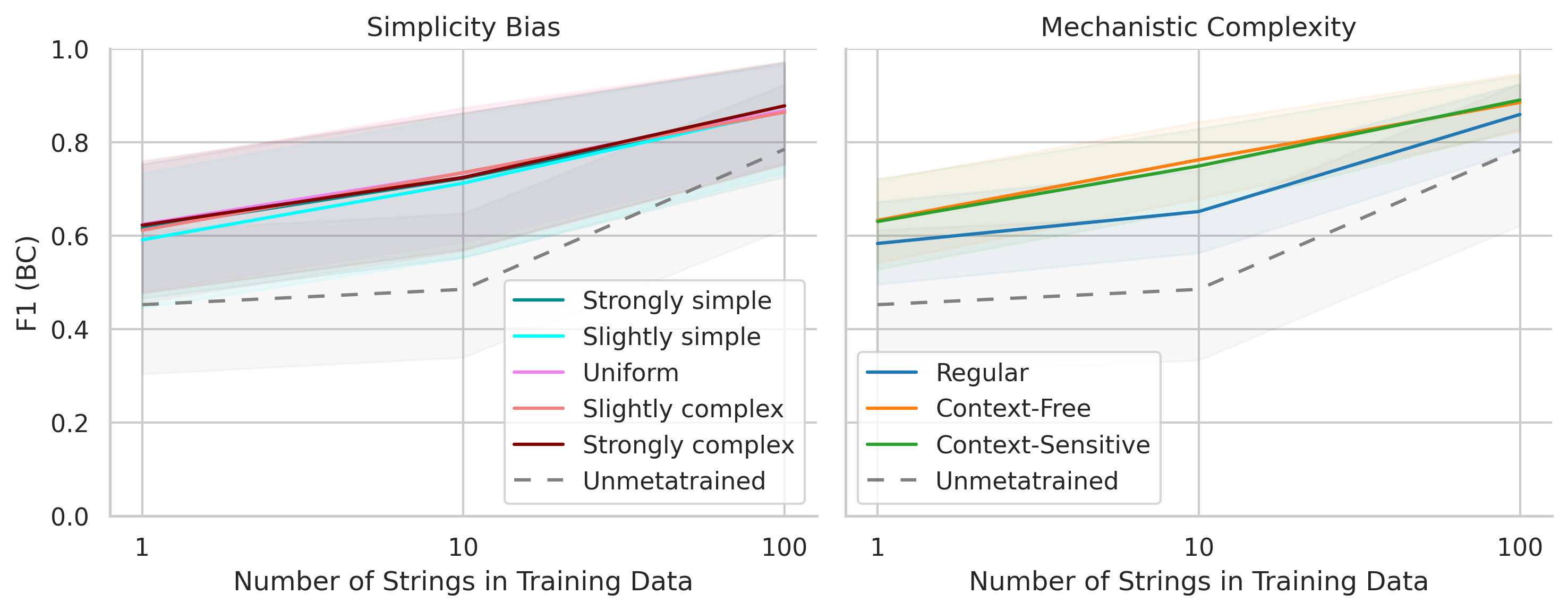}
    \caption{F1 calculated with \textsc{BC}}
  \end{subfigure}
  \begin{subfigure}[t]{\textwidth}
    \centering
    \includegraphics[width=0.9\textwidth]{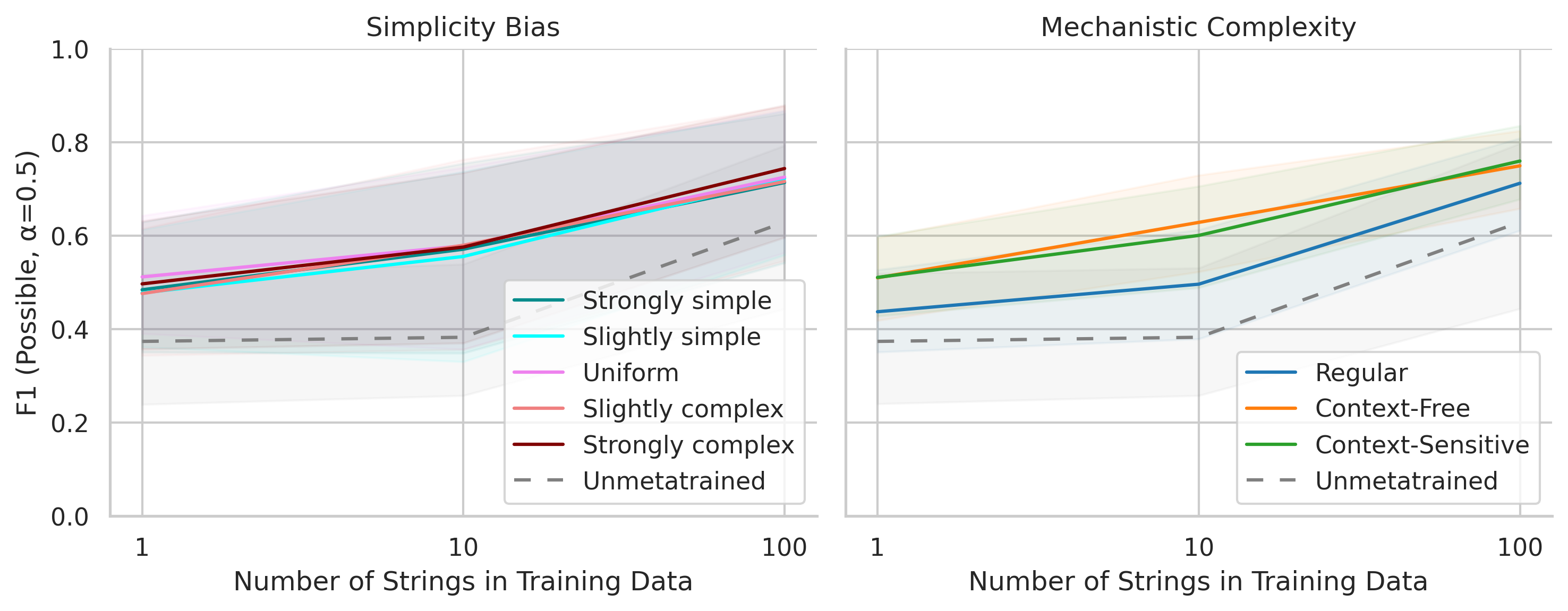}
    \caption{F1 calculated with \textsc{possible} with $\alpha=0.5$}
  \end{subfigure}
  \begin{subfigure}[t]{\textwidth}
    \centering
    \includegraphics[width=0.9\textwidth]{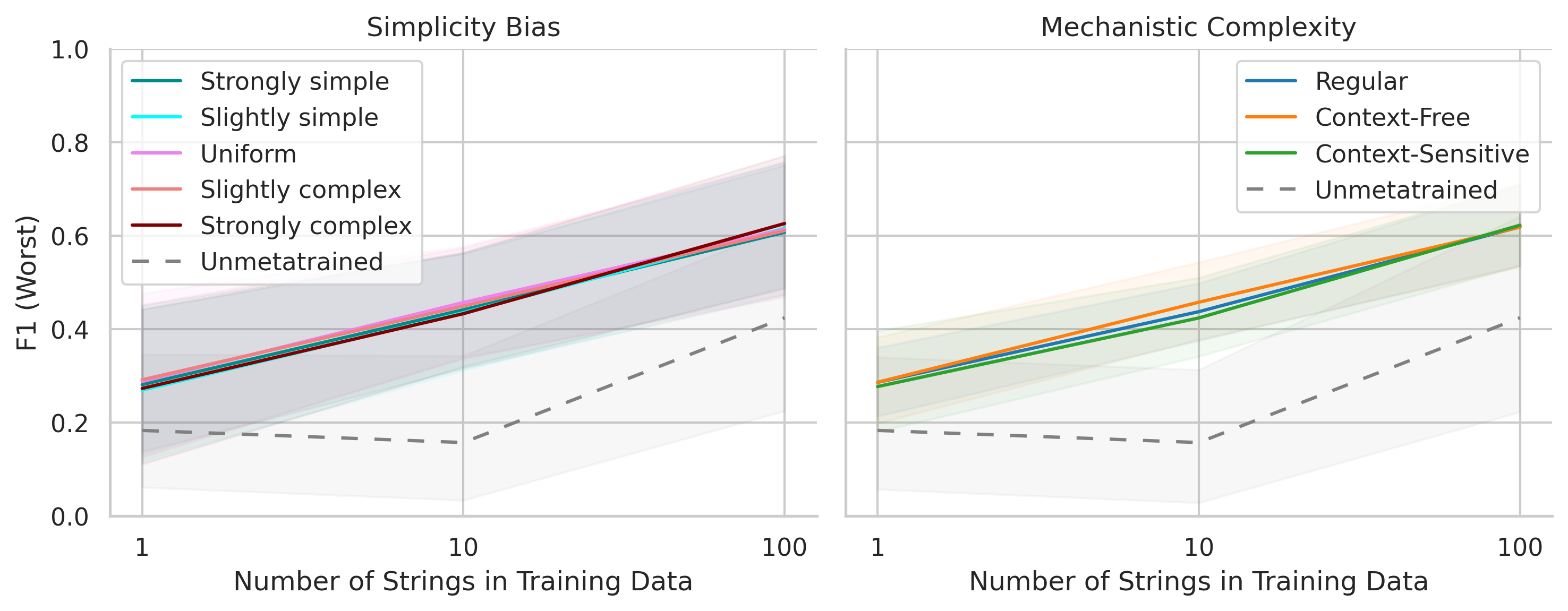}
    \caption{F1 calculated with \textsc{Worst}}
  \end{subfigure}

  \caption{\label{fig:variantplot}
  Results for the variant metrics defined in Figure~\ref{fig:variant}}
\end{figure*}

\end{document}